# MMASD+: A Novel Dataset for Privacy-Preserving Behavior Analysis of Children with Autism Spectrum Disorder

**Pavan Uttej Ravva**[1], **Behdokht Kiafar**[1], **Pinar Kullu**[1], **Jicheng Li**[1], **Anjana Bhat**[1], **Roghayeh Leila Barmaki**[1]

[1]University of Delaware
ravva@udel.edu, kiafar@udel.edu, lijichen@udel.edu, abhat@udel.edu, rlb@udel.edu

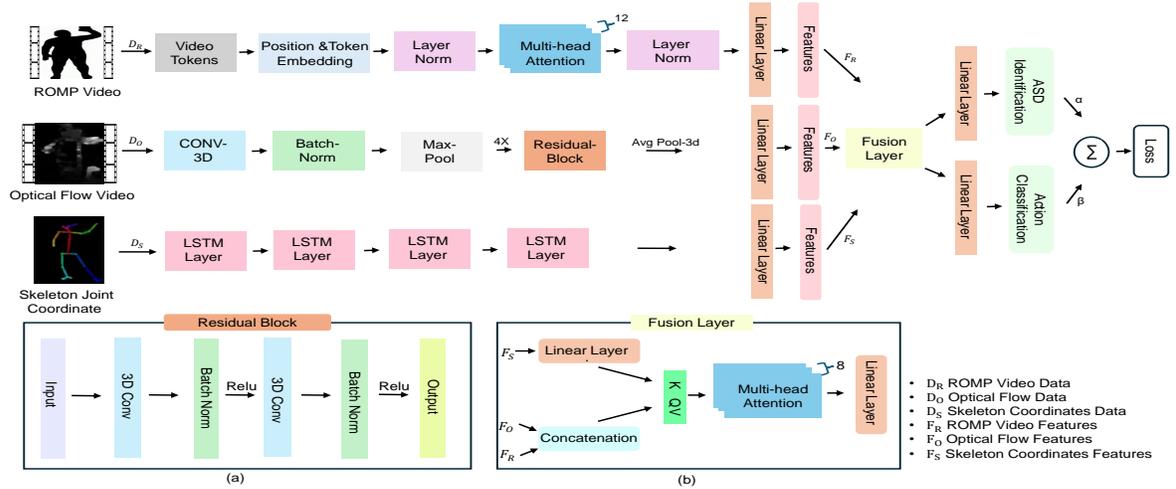

Figure 1: Multimodal Transformer framework for Action and ASD Classification: We integrate ViViT for ROMP Video, 3D-CNN for Optical Flow, and LSTM for skeleton coordinates. A fusion layer with Multi-Head attention combines these features for accurate action and ASD detection using the MMASD+ dataset.

## Abstract

Autism spectrum disorder (ASD) is characterized by significant challenges in social interaction and comprehending communication signals. Recently, therapeutic interventions for ASD have increasingly utilized Deep learning powered-computer vision techniques to monitor individual progress over time. These models are trained on private, non-public datasets from the autism community, creating challenges in comparing results across different models due to privacy-preserving data-sharing issues. This work introduces *MMASD+*[1], an enhanced version of the novel open-source dataset called Multimodal ASD (*MMASD*). *MMASD+* consists of diverse data modalities, including 3D-Skeleton, 3D Body Mesh, and Optical Flow data. It integrates the capabilities of Yolov8 and Deep SORT algorithms to distinguish between the therapist and children, addressing a significant barrier in the original dataset. Additionally, a Multimodal Transformer framework is proposed to predict 11 action types and the presence of ASD. This framework achieves an accuracy of 95.03% for predicting action types and 96.42% for predicting ASD presence, demonstrating over a 10% improvement compared to models trained on single data modalities. These findings highlight the advantages of integrating multiple data modalities within the Multimodal Transformer framework.

## Introduction

Autism spectrum disorder (ASD) is a neurological condition that leads to significant issues in social communication, along with challenges in understanding and expressing communication cues. In the United States, it is estimated that over 1 million children are affected by ASD (Centers for Disease Control and Prevention 2023).

In recent years, deep learning techniques have been extensively applied to analyze autism (De Belen et al. 2020), reducing the need for human experts to manually analyze therapy sessions. These techniques have been instrumental in tasks such as autism diagnosis (Wall et al. 2012), emotion recognition (Marinoiu et al. 2018), and movement pattern assessment (Li, Bhat, and Barmaki 2021).

A common practice among Artificial Intelligence (AI) and Machine Learning (ML) researchers is to compare the per-formance of the proposed model with benchmark datasets. However, in the field of autism research, the availability of open-source benchmark datasets is limited due to privacy concerns, lack of proper labels, and scarcity of clean data. These factors pose significant challenges for analyzing model performance, highlighting the need for labeled, open-source, privacy-preserved datasets.

---
[1]The data and code generated during this study are publicly accessible via https://github.com/pavanravva/Enhanced-MMASD.

To address these challenges, Li et al. (2023) introduced the novel open-source dataset called Multimodal ASD (MMASD).MMASD provides a foundation for research with multiple data modalities, facilitating the analysis of therapeutic sessions. However, the original MMASD dataset presents certain limitations, particularly the inability to distinguish between therapists and children during therapy sessions, which is crucial for accurate analysis. To overcome this, we introduce MMASD+, an enhanced version of the MMASD dataset.

The main goals of MMASD+ are to provide proper labeling, preserve privacy and capture the movement characteristics of individuals during therapy sessions. The MMASD+ dataset includes Optical Flow, 3-Dimensional (3D) Skeleton Coordinates, and complete 3D Body Mesh data extracted from the raw video footage of therapy sessions. This extraction process avoids the exposure of sensitive information and identities of individuals. Therapists and researchers can use this data to gain a deeper understanding of cognitive development in children, monitor therapy progress, and devise personalized treatment strategies.

In this paper, we describe the MMASD+ dataset and its potential application in autism research. Furthermore, we introduce a Multimodal Transformer framework designed to train on MMASD+ data. To evaluate the effectiveness of our framework, we compared its performance against baseline transformer models for computer vision tasks trained on non-privacy-preserving data (video recording of therapy sessions). For comparison we have employed three distinct transformer models: Video Vision Transformers (ViViT) (Arnab et al. 2021), Video Masked Auto Encoder (VMAE) (Tong et al. 2022) and TimeSformer (Bertasius, Wang, and Torresani 2021).

Our Contributions are as follows:

- Enhancement of one of the largest Multimodal Datasets for ASD MMASD to MMASD+: We have developed MMASD+, which improves upon MMASD by integrating a diverse range of data modalities, including Optical Flow, 3D-Skeleton Coordinates, and 3D Body Mesh data, shared with the research community. Additionally, MMASD+ incorporates advanced algorithms to differentiate between therapists and children, enhancing the dataset's utility for accurate analysis.

- Development of a Multimodal Transformer framework: We introduce an open-source Multimodal Transformer framework for action classification and ASD identification. This framework leverages the MMASD+ data, integrating Video Transformers, 3D Convolutional Neural Network (3D-CNN), and Long Short-Term Memory (LSTM) networks for effective analysis.

## Related Work

### Machine Learning Use Cases of ASD

Given the high prevalence of ASD, researchers have increasingly turned to machine learning techniques instead of traditional statistical methods to analyze data (Hyde et al. 2019). Abbas et al. (2018) developed a screening tool for early detection of autism by integrating two separate Random Forest classifiers based on a parental questionnaire and behaviors observed in home videos. Song et al. (2022) proposed an ML-powered diagnostic model, trained using logistic regression, support vector machine, random forest, and XGBoost, to detect comorbid intellectual disability in children with ASD based on behavioral observation data. In another study involving 95 children, Ko et al. (2023) introduced a deep learning model based on joint attention that integrated a CNN, LSTM, and an attention mechanism to differentiate between children with ASD and those with typical development with 97.6% accuracy.

### Existing Datasets of ASD and Privacy-Preservation Considerations

Exploring and analyzing the wide range of behaviors associated with ASD needs multifarious datasets. However the availability of open-source datasets for ASD studies is limited. This creates a demand for open-source dataset for improving and advancing the research in ASD interventions.

Marinoiu et al. (2018) introduced a dataset on humanoid-robot interacting with kids affected by ASD to analyze body movements, facial expressions, and emotional behavior. Similarly, Billing et al. (2020) developed a dataset for robot-enhanced therapy sessions, which collected information on body motion, eye gaze, and head coordinates. Duan et al. (2019) Provided an eye movement dataset, facilitating understanding of the relationship between ASD and eye movement patterns. Additionally, a video dataset has been introduced for computer vision tasks, such as human action recognition and identifying self-stimulatory behavior (Rajagopalan, Dhall, and Goecke 2013). However, these publicly accessible datasets raise significant privacy concerns as they can reveal the identities of the individuals involved.

To tackle this issue, various privacy-preserving frameworks have been proposed. Wang et al. (2024) proposed a deep learning framework for diagnosing ASD using federated learning (FL), designed to train local models without sharing sensitive data, thereby protecting individuals privacy. Rahman et al. (2024) introduced a novel method using an Enhanced Combined Particle Swarm Optimization-Grey Wolf Optimization (PSO-GWO) framework for sanitizing and restoring sensitive ASD datasets, while maintaining privacy and security. Li et al. (2023) also introduced a privacy-preserving dataset from 100 hours of play therapy interventions for children with ASD, incorporating Optical Flow, 2D/3D skeleton data, and clinical evaluations. The major limitation with this dataset is its lack of proper labeling, which does not distinguish between ASD and non-ASD individuals.

To address this challenge, raw video data from the MMASD study (Li et al. 2023) was requested and then pro-cessed to create the MMASD+ dataset. The MMASD+ not only preserves privacy but also provides an improved version of labeled data. In addition, it includes detailed integration of 3D body mesh, providing richer and more precise data for a comprehensive analysis of social interactions and motor activities in children with ASD. In the following, we delve into how Deep Learning techniques are applied to analyze video datasets, particularly through advanced video classification methods.

## Advancing Video Classification with Deep Learning Techniques

The increasing demand for CNNs and their applications has motivated researchers to evaluate the effectiveness of deep learning techniques in computer vision for video analysis. This section explores cutting-edge deep learning models such as 3D-CNNs, and Transformers for video classification tasks.

**3D-CNN:** Early research on video analysis involved feeding hand-crafted features that represented motion and appearance insights (Laptev 2005). With the growing popularity of CNNs for image analysis, Simonyan and Zisserman (2014) introduced the "two-stream" network, that integrated RGB and Optical Flow data in two separate networks for enhanced classification. However, processing video data frame by frame using 2D ConvNets can be inefficient. To address this, researchers have developed 3D-CNNs that capture spatiotemporal features directly from the entire video clip (Tran et al. 2015). This approach has led to significant improvements in video classification tasks by effectively capturing complex motion patterns and temporal dynamics.

**Vision Transformer:** The ability of Transformers to learn relationships within sequential data through their self-attention mechanism (Vaswani et al. 2017) makes them suitable for handling complex temporal patterns, leading to superior performance compared to other models. Many Transformer architectures have been developed in fields such as Natural Language Processing (Kalyan, Rajasekharan, and Sangeetha 2021), Computer Vision (Wu et al. 2020), and Healthcare (Mayer, Cabrio, and Villata 2020).

In this study, we focus on Vision Transformers. These models leverage the self-attention mechanism to capture the spatial relationship within the image, surpassing CNNs (Dosovitskiy et al. 2020). They can be modified and enhanced for applications in video analysis. Arnab et al. (2021) introduced ViViT, which extracted spatiotemporal tokens from videos. These tokens were then fed into attention layers, enabling the model to learn long-range dependencies. ViViT achieved good results on benchmark datasets by utilizing two encoders to learn spatial and temporal relations within the data. Tong et al. (2022) proposed VMAE, which leverages masked autoencoding approach for enhancing the performance of model, this model masks few portions of video and forces it learn both spatial and temporal relations between the data. The space-time attention mechanism was introduced by Bertasius et al. (2021), which helps to capture both spatial and temporal features in distinct layers and helps to analyze long-range dependencies better.

The research on Transformers that handle Multimodal data for classification tasks is underrepresented. So, in this paper, we introduce a novel Multimodal Transformer framework designed to integrate the power of diverse modalities for classification tasks. By integrating Transformer, 3D-CNN, and LSTM network, our framework effectively captures spatial and temporal relationships within complex data. This framework is designed to work with the custom dataset we have created and can be adapted to other publicly available datasets.

## Materials and Methods

This research is approved by the Institutional Review Board (IRB). In this section we describe the dataset, privacy-feature extraction, data preprocessing methods and our Multimodal Transformer framework.

### Dataset

We used raw video data from the MMASD study, proposed by Li et al. (2023), which captures interactions between children with ASD and therapists during therapy sessions. This dataset includes recordings from 32 children (27 males and 5 females), aged 5 to 12 years, who were recruited through local schools, services, and advocacy groups.

The data was collected during triadic play therapy interventions recorded in home settings, where children, therapists, and an adult model engaged in various activities aimed at improving social and motor skills. The interventions were divided into three themes: Robot, Yoga, and Rhythm. In the Robot theme, the child and therapist mimicked robotic movements. Rhythm sessions involved playing musical instruments, while Yoga sessions focused on physical exercises like stretching, balancing, and twisting. These themes were further subdivided into 11 distinct activity classes, as shown in Table 1.

A total of 1,315 video clips were recorded, each with a resolution of $1920 \times 1080$ pixels and an average length of 180 frames. The entire dataset comprises over 244,000 frames.

| S.NO | Action Class | No. Videos |
|---|---|---|
| 1 | Arm Swing | 105 |
| 2 | Body Swing | 119 |
| 3 | Chest Expansion | 114 |
| 4 | Drumming | 168 |
| 5 | Sing and Clap | 113 |
| 6 | Twist Pose | 120 |
| 7 | Tree Pose | 129 |
| 8 | Frog Pose | 113 |
| 9 | Squat Pose | 101 |
| 10 | Marcas Forward Shaking | 103 |
| 11 | Marcas Shaking | 130 |

Table 1: ASD Action Classes and their size

### Person Segmentation and Extraction

In this subsection, we describe the video segmentation process used to isolate individuals in multi-person therapy session recordings. We used YOLOv8 (Redmon et al. 2016) for

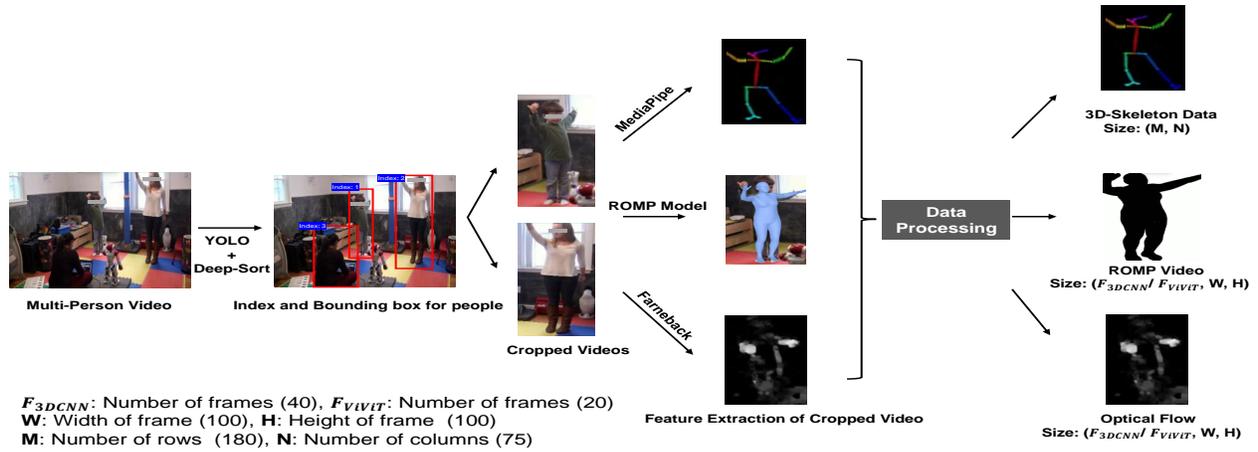

$F_{3DCNN}$: Number of frames (40), $F_{ViViT}$: Number of frames (20)
**W**: Width of frame (100), **H**: Height of frame (100)
**M**: Number of rows (180), **N**: Number of columns (75)

Figure 2: Flowchart illustrating the conversion of raw video recordings into privacy-preserving features for the proposed Multimodal Transformer framework. The process involves detecting and tracking individuals using YOLOv8 and Deep SORT, extracting features with MediaPipe, ROMP, and Farneback algorithms for 3D-Skeleton data, 3D Body mesh, and Optical Flow, and standardizing these features through a data processing pipeline.

detecting individuals in each frame and Deep SORT (Wojke, Bewley, and Paulus 2017) for tracking these individuals across frames.

Initially, the raw videos are resized to meet the input size requirements of the YOLOv8 model. Using pre-trained weights from the Common Objects in Context (COCO) (Lin et al. 2014) dataset, YOLOv8 identifies and outputs bounding box coordinates for individuals in each frame. To ensure that only confident bounding boxes are retained and redundancies are removed, Non-Maximum Suppression is applied.

To track individuals in a video, we extract features from the bounding boxes detected by the YOLOv8 model. These features include textural and appearance characteristics, which are combined to form a unique feature vector for each person. Deep SORT uses these vectors to track individuals across frames, maintaining their identity with the help of Kalman Filters (Li et al. 2015), even as they move.

After extracting the bounding boxes and indices of each person in the video, these data are used to segment individual persons. The area within each person's bounding box is cropped from each frame and compiled to create separate videos for each individual. This process utilizes the bounding box information and indices provided by the YOLOv8 and Deep SORT models. Each video is then manually labeled according to the action class and whether the person has ASD, based on source video verification.

A detailed flowchart illustrating segmentation process, along with subsequent steps of converting the video into privacy-preserving features with pre-processing is shown in Figure 2. A detailed description of privacy-preserving feature extraction and pre-processing steps can be found in the "Privacy-Preserving Feature Extraction" and "Data Processing" sections.

**Privacy-Preserving Feature Extraction**

The segmented individual videos in MMASD+ are utilized to extract privacy-preserving features that retain spatial and temporal information. These videos have an average size of 180 frames. The following features were extracted:

**Optical Flow:** Optical flow data provides information about the direction and velocity of each pixel in a video. We extracted dense Optical Flow data using the Farneback algorithm (Farnebäck 2003). This algorithm approximates the local neighborhood of each pixel with a quadratic polynomial, which helps in calculating motion vector. It also uses a pyramid approach, where Optical Flow is initially calculated at the smallest image scale and then refined at higher levels. This method provides an accurate measurement of the Optical Flow data (Sultana et al. 2022).

**3D-Skeleton:** 3D-Skeleton data provides information about joint coordinates in 3D space. This data is extracted using MediaPipe framework (Lugaresi et al. 2019), which also supports recognition of face meshes and hand gestures. The Mediapipe architecture consists of interconnected nodes known as calculators, which enable feature extraction and image transformation. 3D-Skeleton data is useful in many downsampling tasks like action classification (Aubry et al. 2019), quality analysis of motion (Lei et al. 2020), and behavior understanding (Gesnouin et al. 2020). The output from the MediaPipe model is stored in a .csv format, with each row representing a frame and each column corresponding to the coordinates of skeleton joint.

**3D Body Mesh:** A 3D body mesh is a comprehensive reconstruction of the body in 3D space, capturing both spatial and temporal dynamics of body motion. We utilized the pre-trained, one-stage 3D reconstruction method called Regression of Multiple People (ROMP) (Sun et al. 2021). Unlike traditional models that rely on bounding boxes, ROMP employs a body center heatmap and mesh parameter map to predict body meshes directly from a single RGB image, enhancing robustness, especially in occluded images. ROMP processes the video frame by frame, extracting the 3D Mesh data, which is stored in .npy files along with camera parameters, global body orientation, and 2D pose estimation. The Optical Flow, 3D-Skeleton coordinates, and 3D body mesh data are combined together to form the MMASD+ dataset.

## Multimodal Framework

In this section, we present and discuss our Multimodal Transformer framework, as illustrated in Figure 1. The framework composed of four components:

1. **Video Vision Transformer Network**: A 3D mesh Video Vision Transformer Network.
2. **3D-CNN**: An Optical Flow based 3D Convolution Network.
3. **LSTM**: An LSTM network based on 3D joint coordinates.
4. **Fusion Network and Classification Head**: Enhancing model robustness and classification performance through feature integration.

**Video Vision Transformer Network** This work is inspired by the video vision transformer model, ViViT, introduced by Arnab et al. (2021). The model consists of patch and positional embedding, Multi-Head attention layers, and a classification head. During training, the model learns to generate a latent space vector representation of video data, which is then fed to a fusion layer for classification tasks. A detailed explanation of the layers is provided below.

**Patch and Positional Embedding:** ViViT is a modified version of the Vision Transformer (ViT). While ViT processes image data by dividing it into patches for training, ViViT handles video data, which consists of a sequence of frames (images). These frames are divided into patches, and to capture the temporal relationships between frames, the patches from sequential frames are combined to form a volumetric patch. The number of frames combined in this process is determined by tubelet size $T$.

The dimensions of the video data are represented as $V \in \mathbb{R}^{C \times F \times H \times W}$, where:

- $C$ is the number of channels,
- $F$ is the number of frames,
- $H$ is the height of each frame, and
- $W$ is the width of each frame.

The dimension of volumetric patch representation of video is $V \in \mathbb{R}^{(C \times \frac{F}{T} \times T \times \frac{H}{P} \times \frac{W}{P} \times P \times P)}$, where:

- $T$ is the tubelet size, and
- $P$ is the patch size.

The volumetric patches are passed into the embedding layer. Subsequently, positional information is added to embedded representation by positional embedding layer, resulting in the final size $V \in \mathbb{R}^{(C \times (\frac{F}{T} \times \frac{H}{P} \times \frac{W}{P}) \times (P \times P))}$

**Multi-Head Attention Network:** The Multi-Head attention network helps in understanding the relationships between volumetric patches (Vaswani et al. 2017). This network takes the output of positional embedding and calculates attention weights to establish these relationships. A Multi-Head attention network operates in parallel, with each head having its own set of attention parameters. We have chosen twelve attention heads, the final output is obtained by concatenating the outputs from all heads, which is then passed through a linear layer. The attention is calculated by Equation 1, where:

- $Q$ (Query) is the input query vector,
- $K$ (Key) is the input key vector,
- $V$ (Value) is the input value vector, and
- $d_k$ is the dimensionality of the key vectors.

$$\text{Attention}(Q, K, V) = \text{softmax}\left(\frac{QK^T}{\sqrt{d_k}}\right) V \quad (1)$$

**3D-CNN:** The model employs a 3D-CNN to capture both spatial and temporal features from video data, unlike regular 2D convolution which only captures spatial features (Gopalakrishnan et al. 2024). Initially, the input is fed into 3D-CNN, followed by a Batch Normalization layer and a Max-Pool layer. The features are then passed through four stages of Residual blocks, where each block consists of two 3D convolutional layers, batch normalization layers, and ReLU activation, along with shortcut connections that bypass the convolutional layers to add input directly to output, this four blocks are chosen to maintain the balance between the depth and complexity, by making sure this architecture captures hierarchical and abstract features while avoiding over-fitting. Also, shortcut connections help reducing the vanishing gradient problem. The model then applies global average pooling and a fully connected layer to feed the learned features into the fusion network.

**LSTM:** The LSTM network, derived from the RNN architecture, is designed to handle 3D joint coordinate data by using memory cells to capture long-term dependencies in sequential data (Staudemeyer and Morris 2019). This network comprises four hidden layers, each containing 64 neurons, which helps in capturing complex patterns in the data while maintaining low computational requirements. while we experimented with varying the number of hidden layers, the model with four hidden layers outperformed in terms of performance and computational load. The output from the LSTM layers is then fed into a linear layer, followed by a fusion layer.

**Fusion Network and Classification Head:** The Fusion Attention Layer integrates features from multiple modalities. In this architecture, the final features from the LSTM network are passed to a linear layer, whose outputs serve as the Key (K) for the attention layers, this helps the in integrating the weights with other data modalities. Features from ViViT and 3D-CNN networks are concatenated and used as the Query (Q) and Value (V). We employ an eight multi-head attention mechanism, ensuring effective integration of information from different modalities. The outputs from these attention layers form a unified representation of the multimodal data, which is then processed by a classification head comprising two separate linear layers: one for action prediction and one for ASD identification. This separation enhances performance for both tasks. Additionally, combining the loss function with predefined weights supports balanced training.

### Data Processing

MMASD+ dataset goes under more processing before training with proposed framework.

**Optical Flow:** To ensure consistency, Optical Flow video data are standardized to have an equal number of frames. Since human movements occur at limited speeds, videos often contain redundant frames. Therefore, we extracted 40 frames per video, which adequately covers the entire sequence of actions by a person and helps reducing computational time. Additionally, each video frame was resized to a resolution of $100 \times 100$ pixels.

**3D-Skeleton extracted from MediaPipe:** Each .csv file containing 3D-Skeleton data stores joint coordinates across the frames of the video. Rows represent sequential frame data, while columns represent individual joint coordinates. To standardize the data, the number of rows (frames) in each file is normalized to 180. Files exceeding 180 rows are truncated, whereas shorter files are padded by repeating the initial rows until they reach 180 rows.

**3D Mesh data:** For this process, we converted 3D mesh coordinates into a 3D Mesh video. Each frame file contains 6, 890 3D coordinates representing the entire body. To create a video from these coordinates, we first project them onto a 2D XY plane. Each coordinate is represented as the center of a circle with a fixed radius, and pixels within these circles are darkened to form a pixel-based image. This conversion is applied to all frame files in a video sequence. The resulting images are then compiled to create a video. To ensure consistency, we used a method similar to Optical Flow data processing, extracting 40 frames to cover the entire action sequence.

**Model-Specific Frame Selection:** For data augmentation, the data from the three modalities are rotated by angles of 5, 10, -5, and -10 degrees. To train the 3D-CNN model, all 40 frames are used, ensuring a comprehensive representation of the action sequence. Meanwhile, for the ViViT model, We experimented with both 40-frame and 20-frame video data for training, we observed a slightly higher performance for 40-frame video data, but the computational load and time consumed while training is too high. Therefore, we opted 20 frame configuration. This selection adequately represents the entire 40-frame video while maintaining same level of performance, optimizing computational efficiency and memory usage, ensuring that critical information is retained.

| Data Combination | Name | ML Frameworks |
|---|---|---|
| SCD | A1 | LSTM |
| RVD | A2 | 3D-CNN<br>ViViT |
| OFD | A3 | 3D-CNN<br>ViViT |
| SCD + RVD | A4 | LSTM + 3D-CNN |
| SCD + OFD | A5 | LSTM + 3D-CNN |
| RVD + OFD | A6 | CNN + 3D-CNN<br>ViViT + 3D-CNN |
| SCD + RVD + OFD | A7 | LSTM + 3D-CNN + 3D-CNN<br>LSTM + ViViT + CNN |

Table 2: Data Modalities and ML Frameworks

## Experiments

In this section, we systematically evaluate the performance of our proposed Multimodal Transformer framework, alongside various other computational frameworks, in two distinct experimental settings: (1) independent training for action prediction and ASD identification, and (2) joint prediction of action classes and ASD status using a combined model. The data was split into training and testing sets with an $80\% - 20\%$ ratio.

### Data Modalities and Machine Learning Frameworks

In our experiments, we utilized both the original video recordings of therapy sessions as non-privacy data and extracted privacy-preserving data modalities: 3D-Skeleton Coordinate Data (*SCD*), ROMP Video Data (*RVD*), and Optical Flow Data (*OFD*). We explored different models, including LSTM, 3D-CNN, and ViViT, to evaluate their effectiveness across various data combinations. By integrating these modalities with diverse machine learning frameworks, we aimed to identify the most effective approaches for action prediction and ASD identification. Table 2 provides a detailed overview of the combinations of privacy-preserving data modalities and the machine learning frameworks used in our experiments. Each combination is labeled for easy reference, with the corresponding models specified for each data type.

### Independent Action and ASD Classification

In in this experiment, we trained separate models to classify distinct action classes observed during therapy sessions and to predict whether individuals are affected by ASD. Table 3 presents the results from this independent setup using the various data modalities and models listed in Table 2. The highest accuracy and F1 scores were achieved by our proposed Multimodal Transformer framework that integrates *SCD*, *RVD*, and *OFD* using ViViT for *RVD*, 3D-CNN for *OFD*, and LSTM for *SCD*. This combination significantly outperformed other frameworks. Specifically, the best results for action classification were an accuracy of **96.8%** and an F1 score of **0.97**, while for ASD classification, the framework achieved an accuracy of **95.31%** and an F1 score of **0.9362**. These results highlight the potential of the MMASD+ dataset and the proposed Multimodal Transformer framework that integrates the three modalities achieving the best performance by capturing and understanding subtle body movements over time.

| S.No | Data-Model Combination | Action Classification | | ASD Classification | |
|---|---|---|---|---|---|
| | | Accuracy | F1 | Accuracy | F1 |
| 1 | **A1**: SCD → LSTM | 0.8237 | 0.8100 | 0.8872 | 0.8870 |
| 2 | **A2**: RVD → 3D-CNN | 0.8247 | 0.8245 | 0.8709 | 0.8707 |
| 3 | **A2**: RVD → ViViT | 0.8976 | 0.8836 | 0.9431 | 0.9300 |
| 4 | **A3**: OFD → 3D-CNN | 0.8600 | 0.8558 | 0.9271 | 0.9269 |
| 5 | **A3**: OFD → ViViT | 0.9024 | 0.8924 | 0.9340 | 0.9400 |
| 6 | **A4**: SCD → LSTM, RVD → CNN | 0.9370 | 0.9400 | 0.9434 | 0.9431 |
| 7 | **A5**: SCD → LSTM, OFD → 3D-CNN | 0.9678 | 0.9560 | 0.9487 | 0.9484 |
| 8 | **A6**: RVD → 3D-CNN, OFD → CNN | 0.9671 | 0.9600 | 0.9210 | 0.9220 |
| 9 | **A6**: RVD → ViViT, OFD → CNN | 0.9689 | 0.9687 | 0.9502 | 0.9511 |
| 10 | **A6**: RVD → 3D-CNN, OFD → ViViT | 0.9505 | 0.9502 | 0.9189 | 0.9185 |
| 11 | **A7**: RVD → ViViT, OFD → 3D-CNN, SCD → LSTM | **0.9680** | **0.9700** | **0.9531** | **0.9523** |
| 12 | **A7**: RVD → 3D-CNN, OFD → ViViT, SCD → LSTM | 0.9564 | 0.9500 | 0.9364 | 0.9362 |

Table 3: Performance of Different Data-Model Combinations for Independent Action and ASD Classification.

**Comparative Performance Analysis:** To provide a comprehensive analysis, we compared the performance of our Multimodal Transformer framework, trained on privacy-preserving features, against open-source models like ViViT, VMAE, and TimeSformer, which use cropped video data. The results for both action classification and ASD identification are presented in Table 4. Our framework outperformed all the compared methods, achieving superior accuracy and F1 scores in both tasks. Additionally, our framework leverages privacy-preserving data modalities, unlike other models that depend on cropped video data. This distinction is crucial, as models trained on video data may be less suitable for ASD research due to the scarcity of publicly accessible datasets. By using privacy-preserving data, our framework offers a more practical solution for ASD studies.

| Model | Action Task | ASD Task |
|---|---|---|
| | Accuracy [F1] | Accuracy [F1] |
| **ViViT** | 0.9485 [0.9481] | 0.9424 [0.9423] |
| **TimeSformer** | 0.9051 [0.9064] | 0.9220 [0.9216] |
| **VMAE** | 0.9372 [0.9342] | 0.9299 [0.9298] |
| **Ours** | **0.9680 [0.9700]** | **0.9531 [0.9523]** |

Table 4: Performance of Transformer models on Non-Privacy data

### Joint Action and ASD Prediction Framework

In the second experiment, we selected the best performing models from the independent setup and combined them into a single framework to predict both action classes and ASD status concurrently. Using our proposed Multimodal Transformer framework, as depicted in Figure 1, we fused each features from each network before classification. This joint prediction setup was evaluated using accuracy, F1 score, and confusion matrix. Our framework achieved an accuracy of **95.03%** and an F1 score of **0.9545** for action classification, while for ASD classification, it reached an accuracy of **96.423%** and an F1 score of **0.9642**. The confusion matrix for action classification tasks is shown in Figure 3.

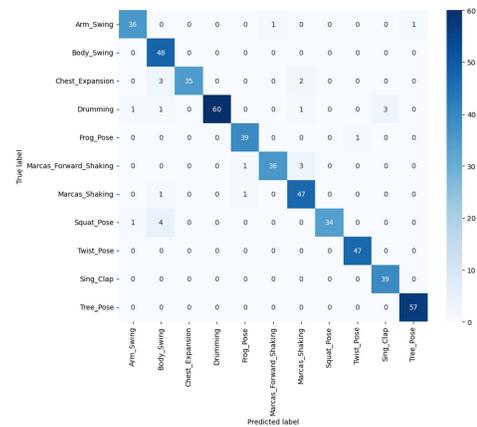

Figure 3: Confusion Matrix of Action Classification

### Conclusion

This study introduces MMASD+, the largest multimodal privacy-preserving dataset focused on children with ASD. MMASD+ includes 3D-Skeleton Coordinates, 3D body mesh, and Optical Flow data, effectively capturing subtle body movements while ensuring the anonymity of the individuals involved. Additionally, we propose a Multimodal Transformer framework that integrates ViViT, 3D-CNN, and LSTM to simultaneously perform action classification and ASD identification, achieving an accuracy of over 95% for both tasks. These results underscore the potential of using our proposed Multimodal Transformer framework to gain valuable insights into the complex dynamics of ASD therapy sessions.